# Approches quantitatives de l'analyse des prédictions en traduction automatique neuronale (TAN)

Maria Zimina[1], Nicolas Ballier[2], Jean-Baptiste Yunès[3]

[1]Université de Paris – maria.zimina@eila.univ-paris-diderot.fr
[2]Université de Paris – nicolas.ballier@u-paris.fr
[3]Université de Paris – jean-baptiste.yunes@u-paris.fr

## Abstract

As part of a larger project on optimal learning conditions in neural machine translation, we investigate characteristic training phases of translation engines. All our experiments are carried out using OpenNMT-Py: the pre-processing step is implemented using the Europarl training corpus and the INTERSECT corpus is used for validation. Longitudinal analyses of training phases suggest that the progression of translations is not always linear. Following the results of textometric explorations, we identify the importance of the phenomena related to chronological progression, in order to map different processes at work in neural machine translation (NMT).

**Keywords:** statistical analysis of textual data, textometrics, neural machine translation.

## Résumé

Dans le cadre d'un projet plus vaste consacré à l'analyse des conditions d'apprentissage optimales pour la traduction, nous cherchons à identifier des phases caractéristiques de l'entraînement des moteurs de traduction neuronaux. Nos expériences ont été réalisées avec OpenNMT-Py. Le pré-traitement a été effectué sur le corpus d'entraînement Europarl et le corpus de validation INTERSECT. Les analyses longitudinales des différentes phases d'entraînement suggèrent que la progression des traductions n'y est pas toujours linéaire. Les analyses textométriques des volets montrent l'importance des phénomènes liés à la progression chronologique et permettent établir progressivement une cartographie des processus à l'œuvre dans la traduction automatique neuronale (TAN).

**Mots clés :** analyses statistiques de données textuelles, textométrie, traduction automatique neuronale.

## 1. Introduction

Le recours à la traduction automatique neuronale (TAN) soulève plusieurs questions concernant l'« apprenabilité » (*learnability*) des réseaux de neurones profonds (RNP). Malgré le succès empirique des RNP, il existe une compréhension théorique limitée des aptitudes d'apprentissage de ces modèles (Zhang *et al*., 2017). Par exemple, une analyse de *learnability* fondée sur la visualisation des réseaux de neurones a donné les premiers résultats dans l'analyse automatique de la reconnaissance des chiffres (Montavon *et al*., 2018). Les cartographies du réseau neuronal décomposées en une série d'opérations géométriques d'étirement et de compression ont également été utilisées pour mesurer le succès empirique des RNP (Li et Ravela, 2019). Malgré ces premiers avancés, la dynamique des réseaux de neurones reste encore relativement mal comprise et beaucoup reste à faire dans la modélisation de l'apprentissage des réseaux, la prédiction et la quantification des incertitudes.

En traduction automatique, la plupart des travaux récents cherchent à améliorer la qualité de la traduction en combinant des moteurs génériques et des moteurs spécialisés pour l'entraînement des modèles (Deng *et al.*, 2017 ; Britz *et al.*, 2019), ou encore en utilisant le contexte étendu (au-delà de la phrase) en langue source et les extensions du contexte bilingue (Tiedemann et Scherrer, 2017 ; Macé et Servan, 2019). Notre travail est plus empirique et s'inscrit dans le cadre d'un projet plus vaste consacré à l'analyse des conditions d'apprentissage optimales pour la traduction. Nous cherchons notamment à identifier des phases caractéristiques de l'entraînement des moteurs de traduction neuronaux en s'appuyant sur l'analyse longitudinale de la progression des résultats d'apprentissage. Appliqué aux sorties des phases d'apprentissage, ce type de recherches permet de comprendre ce qui se passe à chaque étape, qu'il s'agisse de l'identification de problèmes résolus (ou pas), de l'émergence de formes nouvelles (entités nommées, sigles, etc.) ou de propriétés textométriques caractéristiques. Au-delà d'une telle cartographie individuelle des phases, quelle dynamique est-elle à l'oeuvre sur l'ensemble des données produites ? Les expériences présentées dans cette contribution visent à étudier cette question.

## 2. Construction du jeu de données

### 2.1. Caractéristiques de l'entraînement

Nos expériences ont été réalisées avec la version v1.0.0.rc1 d'OpenNMT-Py[1] (Klein et al., 2017), et ses outils onmt_preprocess, onmt_train et ontm_translate. Le pré-traitement a été effectué avec les valeurs par défaut sur le corpus d'entraînement Europarl[2] (bi-texte français-anglais de 2 007 723 phrases) et le corpus de validation INTERSECT[3] (bi-texte français-anglais de 4 967 phrases, Salkie 1995). Les scores BLEU ont été calculés avec sacrebleu 1.4.2 (Post, 2018). Le réseau utilisé est du type LSTM (*long short-term memory*) à 2 couches. La validation (par défaut) est d'une toute les 10 000 étapes. Les calculs ont été menés en 13 heures sur une machine à base de Intel(R) Core(TM) i7-7700K exécutant une distribution Debian GNU/Linux 10 et utilisant une carte GPU NVIDIA GeForce GTX 1080 Ti. L'espace mémoire occupé durant l'entraînement correspond quasiment à l'intégralité de celle disponible sur la carte de calcul, soit environ 11 Go. La traduction a été réalisée avec la graine d'aléa égale à 0 (pour fixer la reproductibilité de l'aléa) en utilisant le processeur graphique (GPU)[4] et sur un corpus de 2 887 phrases (distinct des deux autres corpus). Les paramètres détaillés de l'entraînement sont disponibles sur https://github.com/nballier/NMT/JADT2020.

---

[1] Lancé en décembre 2016 par SYSTRAN et Harvard NLP, OpenNMT est un des plus grands projets Open Source de traduction neuronale avec 18 versions majeures, 1400 branches sur Github et plus de 500 articles publiés (http://opennmt.net/). L'approche de traduction automatique neuronale (TAN) implémentée dans l'outil OpenNMT (Klein *et al.*, 2017) relève de l'approche séquence-vers-séquence (Sutskever *et al.*, 2014) tout en combinant une approche d'attention (Luong *et al.*, 2015).

[2] https://www.statmt.org/europarl/

[3] http://arts.brighton.ac.uk/staff/raf-salkie/intersect

[4] Ce paramètre est optionnel car la traduction ne nécessite pas de puissance de calcul si importante.

*2.2. Données générées*

Le jeu de données obtenu après l'optimisation constitue l'équivalent de 20 époques ou « cycles ». On fait des traductions avec des époques révolues. Une époque *(epoch)* génère une traduction à chaque fois que le corpus d'entraînement a été vu intégralement, que le jeu de données est passé entièrement par l'apprentissage et qu'au moins une rétropropagation a eu lieu, avec une mise à jour du modèle. Pour chaque époque (chaque étape d'entraînement, 01-20), nous disposons également du nombre des mots non traduits *(unknown* : UNK) et du score BLEU[5] calculé pour chaque phrase par rapport au corpus de référence. Les scores BLEU calculés font apparaître des régularités dans les phases d'entraînement : une progression rapide dans les prédictions 1-4 puis de très faibles variations à partir de la 5ème prédiction (cf. Figure 1) :[6]

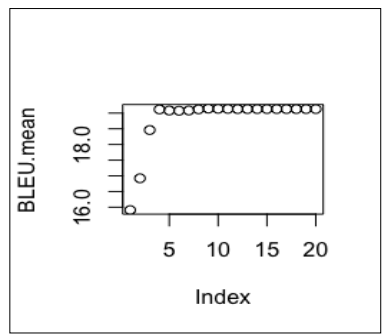

*Figure 1 : Variations des moyennes des scores BLEUS dans les prédictions*

En termes d'analyse de la traduction, les données textuelles générées constituent un corpus parallèle multi-volets composé de 20 versions (prédictions) alignées au niveau de la phrase, complétées par le texte source (79 585 occ.) et la traduction humaine de référence *Target* (66 780 occ.). Seules les **13 premières versions de traduction 01-13 comportent des différences**, les prédictions **14-20 étant des versions identiques** (avec l'augmentation du nombre d'époques et l'allongement de la durée de l'apprentissage la traduction ne progressait plus). Ce corpus pluritextuel appelé ***EPOCHS*** a été transformé en une base textométrique avec un système de parties sur le modèle de données *Trame/Cadre* implémenté dans le logiciel de l'analyse textométrique *Le Trameur* (Fleury et Zimina, 2014) et *iTrameur* (Zimina et Fleury, 2018). La Figure 2 résume les principales caractéristiques lexicométriques 01-13 d'***EPOCHS*** calculées par *Le Trameur* : position sur la *Trame* (début/fin), nombre de formes (fq items), nombres d'occurrences (FQ items), fréquence maximale (Fq Max) et la forme de la fréquence maximale (Forme Max). On remarque notamment que le volet 01 constitue la traduction la plus longue en nombre d'occurrences qui dépasse largement la *Target* :

---

[5] Le score BLEU (*bilingual evaluation under study*) est un algorithme d'évaluation de la qualité de la traduction automatique (Papineni *et al.*, 2002). La qualité est considérée comme la correspondance entre la production d'une machine et celle d'un humain (Post, 2018).

[6] Pour effectuer ces analyses, nous avons utilisé le logiciel R.

| Valeur Partie | début   | fin     | fq item | FQ item | Fq Max | Forme Max |
|---------------|---------|---------|---------|---------|--------|-----------|
| Source        | 1       | 167367  | 7025    | 79585   | 4186   | de        |
| Target        | 167368  | 308882  | 5984    | 66780   | 4350   | the       |
| 01            | 308883  | 462995  | 3572    | 74868   | 7470   | the       |
| 02            | 462996  | 604639  | 3989    | 68428   | 7611   | the       |
| 03            | 604640  | 746502  | 4315    | 68375   | 6099   | the       |
| 04            | 746503  | 888596  | 4351    | 68441   | 6171   | the       |
| 05            | 888597  | 1029879 | 4362    | 68037   | 5904   | the       |
| 06            | 1029880 | 1171142 | 4353    | 68011   | 5930   | the       |
| 07            | 1171143 | 1312325 | 4359    | 67967   | 5927   | the       |
| 08            | 1312326 | 1453605 | 4372    | 68008   | 5935   | the       |
| 09            | 1453606 | 1594842 | 4362    | 67993   | 5933   | the       |
| 10            | 1594843 | 1736053 | 4364    | 67980   | 5922   | the       |
| 11            | 1736054 | 1877245 | 4365    | 67969   | 5918   | the       |
| 12            | 1877246 | 2018449 | 4364    | 67975   | 5916   | the       |
| 13            | 2018450 | 2159650 | 4365    | 67973   | 5916   | the       |

*Figure 2 : Principales caractéristiques lexicométriques des volets **EPOCHS***

## 3. Premières analyses quantitatives

### *3.1. L'accroissement du vocabulaire*

L'analyse de l'accroissement du vocabulaire dans ***EPOCHS*** (cf. Figure 3) montre que toutes les prédictions se distinguent de la traduction de référence où le nombre de formes tend à croître plus rapidement et se rapproche considérablement du texte source en français. Les deux premières prédictions (01 et 02) sont particulièrement éloignées de la *Target*.

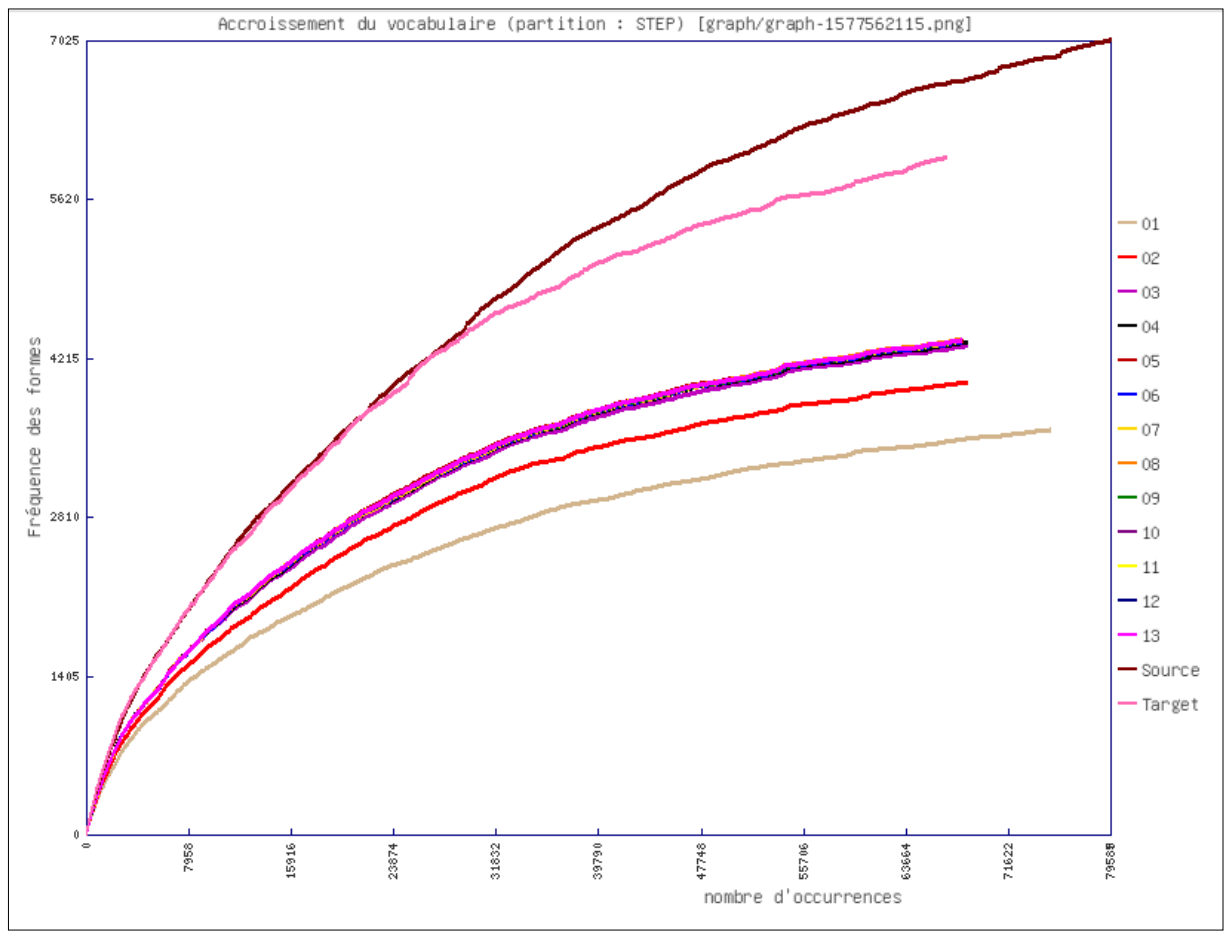

*Figure 3 : L'accroissement du vocabulaire dans les différents volets d'**EPOCHS***

### *3.2. L'analyse des spécificités d'EPOCHS*

Pour tenter de comprendre les différences constatées dans l'accroissement du vocabulaire, nous nous sommes intéressés aux mots caractéristiques des prédictions à l'aide de l'analyse des *spécificités* (Lebart et Salem, 1994). Plusieurs spécificités remarquables (avec un indice de spécificité >10) correspondent à des formes de haute fréquence dont la ventilation est marquée par des différences notables en termes de fréquences d'emploi dans les deux premières prédictions, qui se distinguent nettement des états qui suivent (cf. Figure 4).

La Figure 4 atteste que la 1ère prédiction est caractérisée par un suremploi des propositions subordonnées introduites par *that* et fait largement appel à la préposition *of*, ce qui indique un suremploi des procédés de base utilisés en anglais pour relier deux noms entre eux (N *of* N).

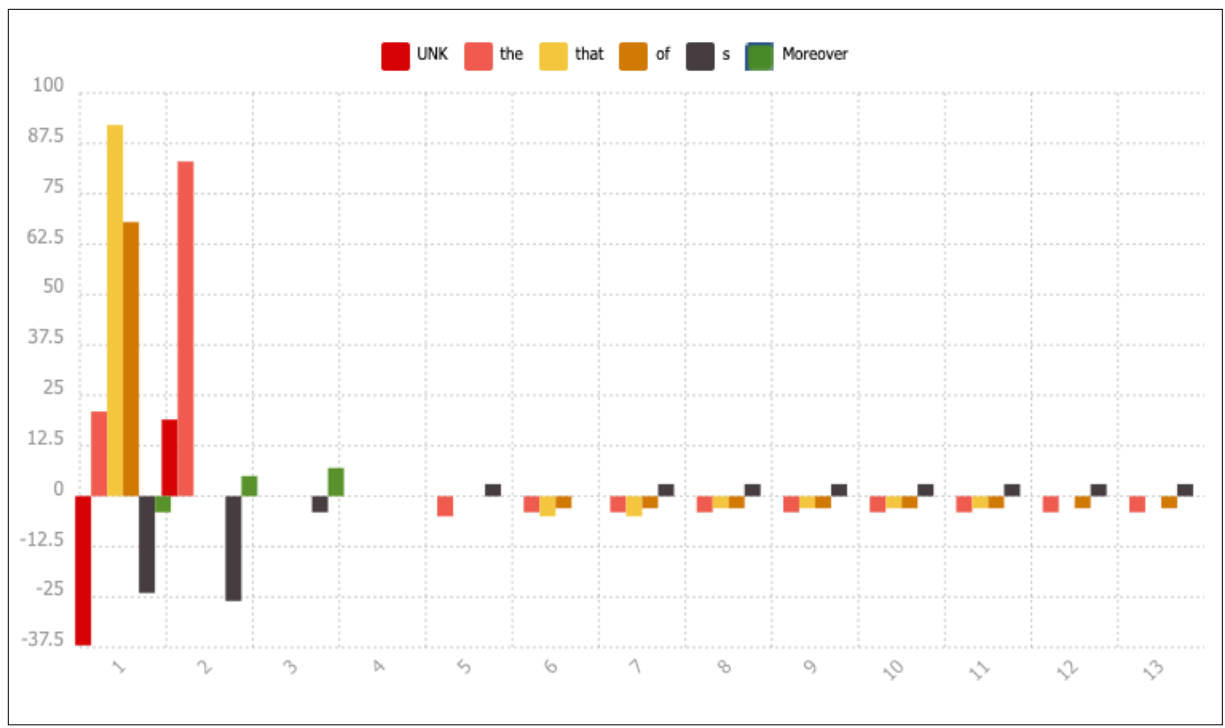

*Figure 4 : L'accroissement du vocabulaire dans **EPOCHS** (1-13)*

Si l'article défini *the* est sur-employé dans la 1ère prédiction[7], c'est surtout dans la 2ème prédiction que sa spécificité est remarquable. Nos analyses ont montré que *the* utilisé dans cette partie rentre dans plusieurs segments répétés comportant des mots non reconnus (*UNK*) dont le nombre augmente significativement dans la 2ème prédiction : la syntaxe des phrases s'affine mais des îlots non traduits sont plus nombreux, comme dans le cas du segment *UNK the UNK* (+E07) : *The strategy should consist of **UNK the UNK** relations by forming special coalitions* /…/ En même temps, les noms à mettre au premier plan ne sont pas encore connus et les séquences N'*s* N sont nettement sous-représentées.

L'absence des N'*s* N est encore plus spécifique dans la 3ème prédiction. En revanche, l'emploi des connecteurs tels que le *Moreover* en début des phrases devient caractéristique dans cette partie du corpus et pourrait indiquer une forme de précision dans la structuration du discours. Cette même tendance progresse en 4ème prédiction où le sous-emploi des N'*s* N diminue et les mots de liaison sont également suremployés.

---

[7] En pratique, notamment lorsque le corpus d'entraînement est très restreint, la répétition des *the the the* est souvent la traduction produite par défaut.

La 5ème prédiction marque une forme de stabilisation. On remarque que les variations de fréquences d'emploi sont beaucoup plus faibles au fur et à mesure des prédictions successives (6-13). A partir de l'état 6, les relations entre les noms semblent mieux maitrisées et la présence des contextes avec N'*s* N devint caractéristique, tandis que les fréquences d'emploi de *the, that* et *of* basculent en spécificités négatives. Les mots non traduits *UNK* ne déclenchent aucune spécificité dans cette dernière série de prédictions.

## 4. Typologies des époques d'entraînement

### *4.1. Classification des époques par nombre de mots non-reconnus*

Nous avons d'abord utilisé l'algorithme VNC : *variability-based neighbour clustering* (Gries et Hilpert, 2008) afin de réaliser une classification des phases sur la base des mots non-reconnus. La méthode VNC sert à produire une périodisation motivée statistiquement et tient compte de la structure diachronique des données (la mesure de similarité se fait par voisinage, non dans l'absolu). Les vecteurs correspondant aux nombres de mots non-reconnus, pour les vecteurs de phrases traduites à chaque époque, sont évalués à partir de la mesure de similarité donnée par le coefficient de Pearson (*Pearson product-moment correlation coefficient*). L'intérêt de l'approche réside dans sa capacité à produire des comparaisons adjacentes terme à terme et de forcer en quelque sorte le dendrogramme à s'organiser sur la chronologie (ici, des phases d'entraînements).

Pour les données *EPOCHS*, l'algorithme VNC montre que les changements qui s'opèrent des époques 6 à 20 sont mineures, la distance y est bien moindre qu'entre les époques 1 et 5 (cf. Figure 5).

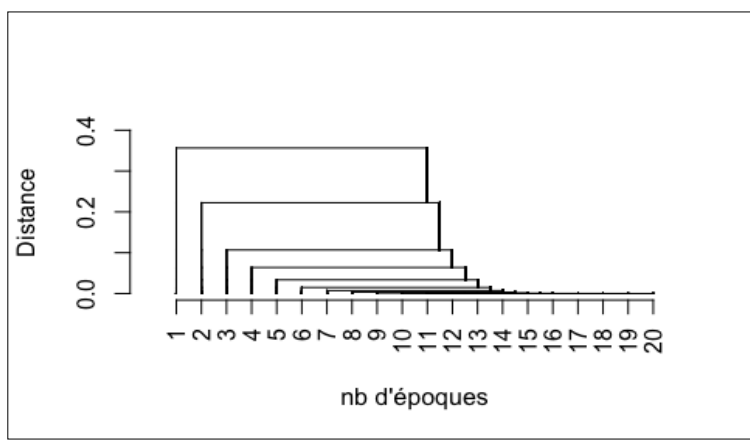

*Figure 5 : Clustering par voisinage des époques à l'aide de l'algorithme VNC (Gries et Hilpert, 2008)*

### *4.1. Analyse longitudinale des époques par nombre de phrases sans segments inconnus*

Dans la classification des époques par nombre de mots non-reconnus, nous avons retenu un algorithme exploité pour la périodisation en diachronie, qui « force » l'interprétation vers la dernière époque. Le suivi longitudinal des segments non-reconnus constitue un autre critère typologique permettant une périodisation de l'apprentissage. Dans ce cas, la recherche d'une possible phase de convergence optimale pourrait donner des résultats différents de ceux obtenus en prenant en compte uniquement le score BLEU. Nos expériences ont montré que, si l'on retient comme critère le nombre de segments non-reconnus dans les époques, la convergence ne serait pas optimale en phase 13 mais plutôt en phase 10 (cf. Figure 6).

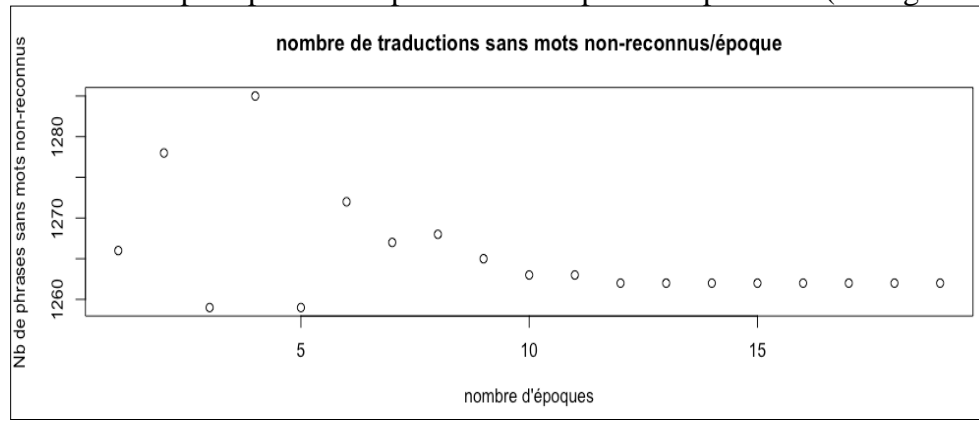

*Figure 6 : Nombre de traductions sans mots inconnus dans EPOCHS*

### 4.3. Analyse factorielle des correspondances du corpus EPOCHS

Les premières analyses quantitatives d'***EPOCHS*** ont montré l'importance des phénomènes liés à la progression textuelle au fil de l'apprentissage. Pour arriver à une représentation typologique de la progression chronologique des époques (prédictions) nous nous appuyons sur les résultats de l'analyse factorielle des correspondances (Lebart et Salem, 1994). La visualisation fournie par l'analyse factorielle (cf. Figure 7) souligne la structure chronologique de la première série de l'entraînement (01-5) avec une stabilisation successive à partir de l'époque 6 qui peut être apparentée à une phase de « révision » (06-13). Les prédictions 13-20 étant strictement identiques, elles sont exclues de cette analyse.

Sur la Figure 7, les traits rouges qui relient les parties consécutives sont rajoutés pour mettre en évidence le résultat fournis par l'AFC : une évolution progressive du vocabulaire au fil des parties et un système de distances sur l'ensemble des parties. La décomposition de ces distances selon les différents axes de la représentation factorielle montre notamment que le premier facteur rend compte d'une évolution linéaire. Nous constatons ainsi que la progression chronologique des époques confère aux données des propriétés particulières qui peuvent les apparenter aux séries textuelles chronologiques (Salem, 1991) dans la première phase d'entraînement (01-05), suivie d'une série de révisions dans la deuxième (06-13) et une stabilisation complète (prédictions identiques) dans la troisième (13-20).

### 4.4. Comparaison avec Target

Sur la Figure 8, l'analyse des correspondances intègre la traduction de référence *Target*. Les résultats de l'AFC montrent que la traduction de référence se trouve à l'origine des axes, avoisinant les séries 06-13, nettement opposées aux prédictions de départ 01-02 (cf. Figure 7). La distance entre la traduction de référence et les prédictions peut apporter un éclairage supplémentaire sur la progression de l'apprentissage mais ne constitue en aucun cas un paramètre exclusif de notre analyse.

Dans la continuité de cette exploration, l'analyse des spécificités de la traduction de référence fait émerger une liste d'entités nommées, sigles, variantes orthographiques, etc., qui distinguent la traduction humaine de la traduction automatique.

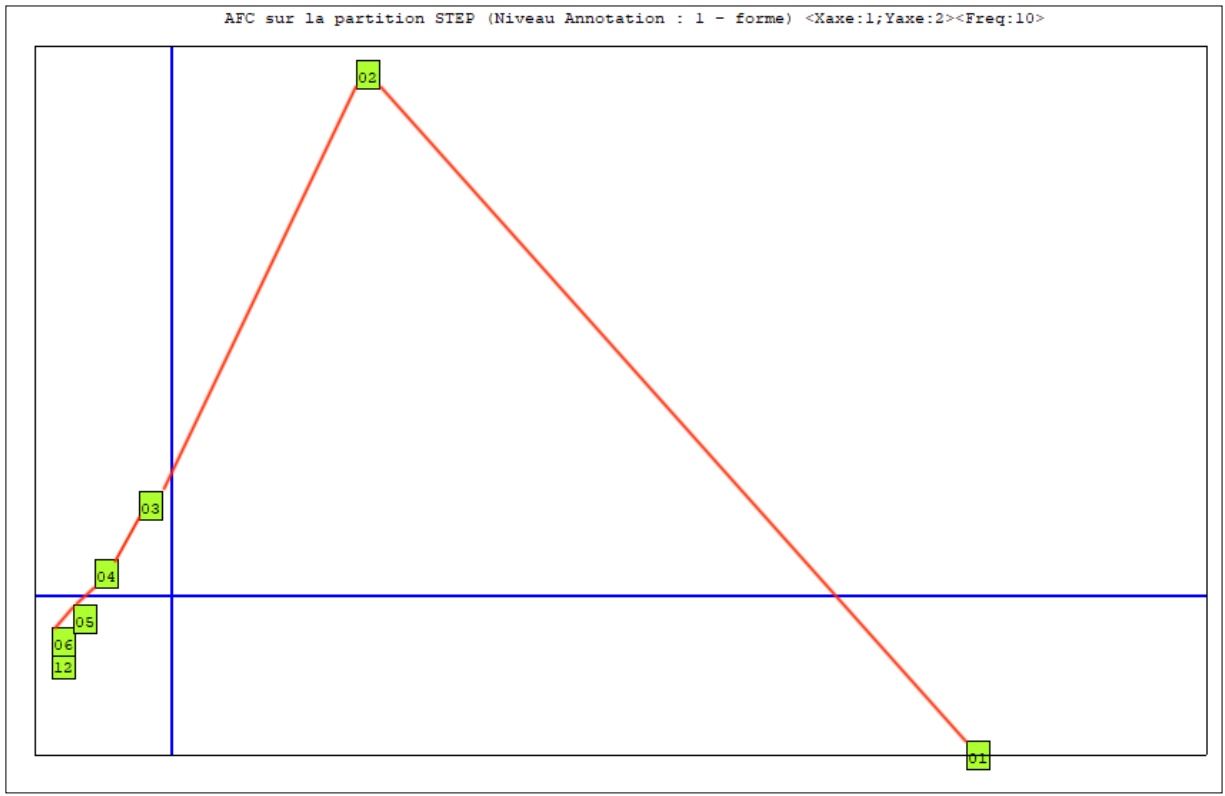

*Figure 7 : Analyse factorielle des correspondances EPOCHS 01-13
(dim 1 : 37%, dim 2 : 23 %)*

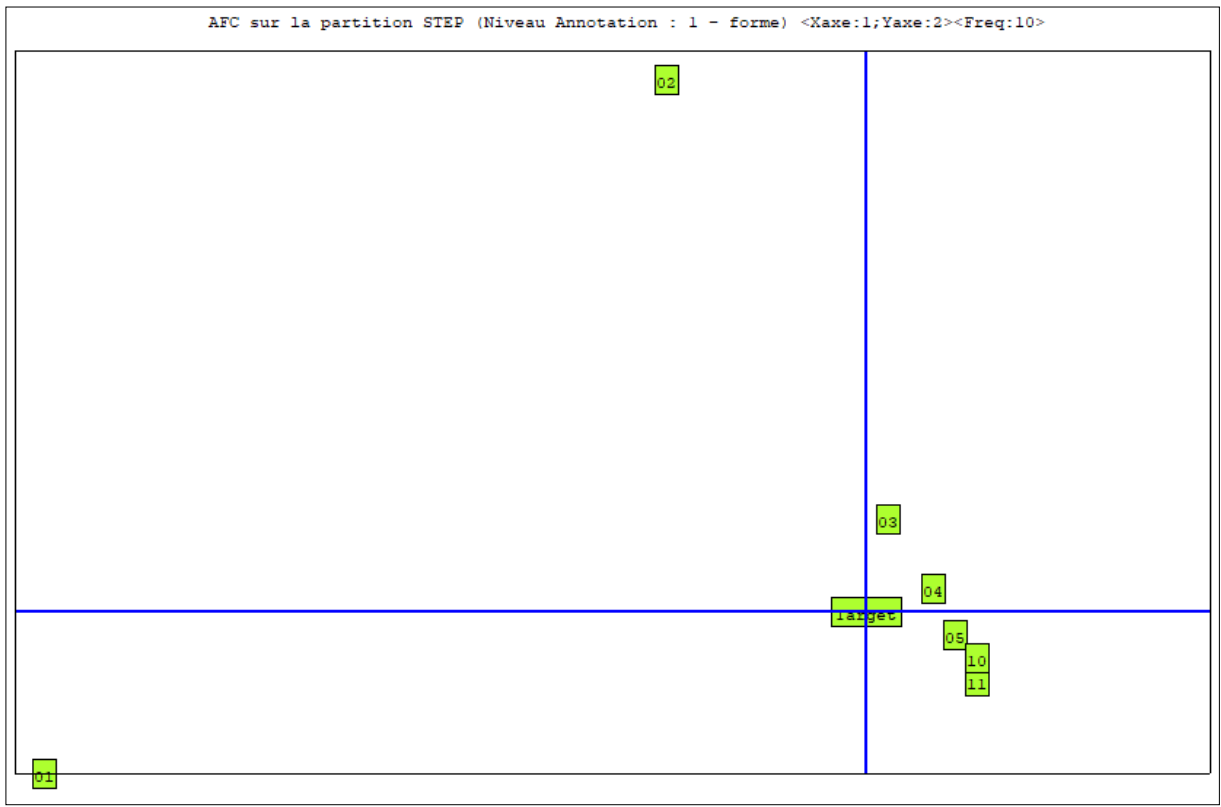

*Figure 8 : Analyse factorielle des correspondances EPOCHS 01-13 et Target
(dim 1 : 37%, dim 2 : 23 %)*

La Figure 9 résume quelques exemples de spécificités remarquables. Pour chaque mot, les scores au-dessus de l'arc indiquent (dans l'ordre) : l'indice de spécificité (supérieur à +40), la fréquence absolue dans le corpus, et la fréquence dans la partie. Les constats de l'analyse des spécificités peuvent mener à une construction semi-automatique de dictionnaires associés au modèle de traduction (des fonctions sont prévues à cet effet dans OpenNMT) pour recenser des éléments qui posent problème lors de l'apprentissage.

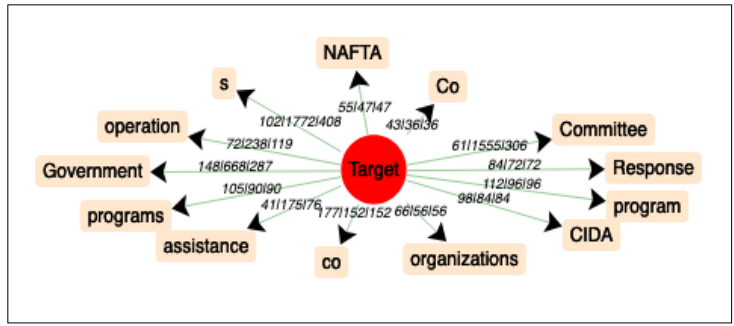

*Figure 9 : iTrameur. Spécificités de la traduction de référence Target −IndiceSpécif|FQ|fq→mot*

## 5. Lecture textométrique différentielle d'EPOCHS

### *5.1. Principes de navigation*

Les analyses typologiques ont permis de converger vers une première compréhension des phases d'apprentissage. Une cartographie plus détaillée des processus à l'œuvre dans la TAN s'appuie sur une intervention interprétative contextualisée. Amorcée par les résultats quantitatifs, elle mobilise la *lecture textométrique différentielle* (LTD). La LTD montre au fil de la lecture comparative de textes leurs différences, ou leurs similarités caractéristiques (Gledhill *et al*., 2017). La Figure 10 permet de comparer à l'aide des segments répétés les prédictions et la traduction de référence (présentée en dernier). Les segments répétés sont matérialisés par des soulignements. Les évolutions observées dans les phases de la TAN mises en correspondance avec la traduction humaine fournissent des indices sur des éléments « difficiles » à appréhender dans le cadre expérimental fixé au départ. On remarque, par exemple, que l'expression « brouiller les cartes » issue du contexte original « Le contexte mondial en évolution La fin de la guerre froide a brouillé les cartes en matière de sécurité, non seulement en Europe mais dans le monde entier » pose plusieurs problèmes en TAN. Parallèlement, l'entité nommée « la guerre froide » n'est pas complètement reconnue dans les premières prédictions :

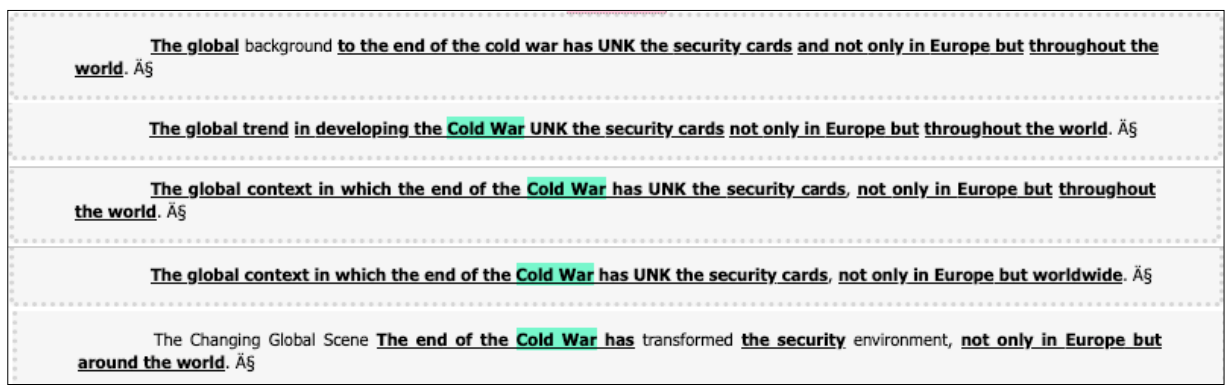

*Figure 10 : Comparaison des prédictions avec la traduction de référence (iTrameur)*

Les contextes présentés sur la Figure 10 invitent à une analyse systématique des entités nommées, qui pourrait s'appuyer sur l'émergence de majuscules ou d'autres signes typographiques dans les prédictions successives.

### 5.2. Recherche des amorces par Barycentre temporel / Coefficient Von Neumann

Nous avons vu que les **EPOCHS** ont des caractéristiques des séries textuelles chronologiques. Pour amorcer l'intervention interprétative, nous nous appuyons sur la périodisation *Barycentre Temporel (BT) / Coefficient Von Neumann (VN)* (Salem, 1991). Le calcul permet de mettre en avant des formes qui évoluent dans le temps en tenant compte de la structure chronologique des données. Le BT mesure la répartition d'une forme : on mesure à quel endroit de la partition on a autant d'occurrences avant qu'après. L'indice de VN mesure les écarts de fréquence d'une forme entre les différentes parties : si les écarts sont stables, la forme évolue de manière continue tout au long de la chronologie ; si la fréquence varie beaucoup d'une partie à l'autre, on repère un profil « accidenté » avec des hauts et des bas. Enfin, la combinaison de ces deux indications met en avant les formes qui évoluent en fonction de la chronologie (on trie par BT et on repère des VN faibles). Les résultats calculés par *iTrameur* sont éclairants car ils permettent de répertorier des changements contextuels qui interviennent à partir de l'époque 5 : ils sont plus difficiles à capter par des seuils de spécificité car les écarts de fréquences sont très faibles (cf. *Figure 11*). Dans le volet 8, par exemple, la forme *Nordic* apparaît dans le groupe nominal *Nordic population* en remplacement de *northern* :

| Item | FQ | BT | VN | 5/fq | 6/fq | 7/fq | 8/fq | 9/fq | 10/fq | 11/fq | 12/fq | 13/fq |
|---|---|---|---|---|---|---|---|---|---|---|---|---|
| vis | 12 | 6.5 | 0.25 | 0 | 0 | 0 | 2 | 2 | 2 | 2 | 2 | 2 |
| Nordic | 6 | 6.5 | 0.25 | 0 | 0 | 0 | 1 | 1 | 1 | 1 | 1 | 1 |
| Table | 6 | 6.5 | 0.25 | 0 | 0 | 0 | 1 | 1 | 1 | 1 | 1 | 1 |
| à | 6 | 6.5 | 0.25 | 0 | 0 | 0 | 1 | 1 | 1 | 1 | 1 | 1 |

*Figure 11 : LTD : recherche des amorces par Barycentre temporel / Coefficient Von Neumann*

*L'extrait du tableau généré par iTrameur donne à voir pour tous les mots de fréquence générale supérieure à la valeur **FQ MAX** (> 5) d'une part leur fréquence absolue (**FQ**) sur toutes les parties induites par la partition sélectionnée (**EPOCHS**) puis un indicateur appelé barycentre temporel (**BT**), le coefficient de Von Neumann (**VN**) et enfin les fréquences (**fq**) dans les volets 5-13 d'**EPOCHS**.*

## 6. Résultats et discussion

### 6.1. Périodisation de l'apprentissage

Plusieurs méthodes quantitatives convergent pour suggérer une caractérisation similaire de la périodisation des époques. Après une phase de transformations importantes jusqu'à l'époque 5, des transformations d'importance moindre se manifestent jusqu'à l'époque 13. La classification/structuration se stabilise ensuite. Naturellement, ces observations concernent les corpus étudiés et pourraient être sensiblement différentes dans le cas des langues éloignées et/ou avec d'autres types de données textuelles de départ. Une expérience nous a montré que plus rien ne changeait après l'époque 13 si on portait le nombre d'époques à 60.[8] Au plan interprétatif, une division du travail semble apparaître selon les époques. De l'époque 1 à l'époque 5, le moteur semble apprendre essentiellement la structure, les mises en correspondances plus fines intervenant entre les époques 5-10 : on constate l'apparition des connecteurs, des entités nommées, une diversification du lexique. Les révisions qui s'opèrent vers la fin, entre les époques 10-13, sont marginales et correspondent souvent à des paraphrases, plus ou moins justifiées.

*6.2. Cartographie des révisions*

Une approche topographique implémentée dans les outils de textométrie (Zimina et Fleury, 2018) permet de localiser avec précision les contextes correspondant aux profils étudiés sur une carte des sections parallèles pour faciliter l'interprétation des révisions contextuelles identifiés par la LTD, telles que l'apparition de l'unité *vis-à-vis* diagnostiquée par la périodisation sur la Figure 11 :

**It is not directly directed at parliaments**, *and even less towards the citizens, that their duty is that obligation (époque 5).*

**It is not directly vis-à-vis parliaments**, *and even less towards the citizens, that their duty is that obligation (époques 8-13).*

## 7. Conclusion et perspectives

Dans une perspective idéale, il s'agit d'établir une cartographie des processus à l'œuvre dans la TAN (dont rien n'assure qu'elle soit vraiment comparable aux pratiques des traducteurs). Nous voudrions disposer d'un tableau analytique qui, d'un point de vue linguistique, pose l'ensemble des transformations observables d'une époque à une autre, qui tente de caractériser les époques à partir de l'émergence de phénomènes (reconnaissance des entités nommées, élucidation des segments présentés comme non-reconnus aux phases précédentes, etc.). Par exemple, les contextes présentés sur la Figure 10 invitent à une analyse systématique des entités nommées afin de se prononcer sur les différences éventuelles d'identification entre procédure humaine et automatique pour l'association des séquences en majuscules à une présomption d'entité nommée reconnue. La détection automatique des entités nommées et leur traduction neuronale fait partie des projets en cours.

*7.1 Du suivi des sorties au suivi des entrées*

Notre analyse de la *learnability* se porte sur trois fronts, que l'on peut rapporter aux trois temps *input, process, output*. Le présent travail contribue à l'analyse des sorties successives pendant la phase d'entraînement. Un autre travail en cours (Ballier *et al*., 2020) porte sur l'impact des annotations en entrée sur les traductions. Enfin, pour l'analyse des processus, nous souhaitons utiliser les techniques de visualisation de l'activité des réseaux de neurones (Montavon, 2018) et étudier l'influence de différents paramètres sur les traductions. Parmi les paramétrages de l'entraînement pourront être envisagés la taille du mini-lot (*mini-batch*), le type de régularisation, le taux d'apprentissage, le facteur de dégradation et le cycle auquel il s'applique (cf. les paramétrages pour les modèles génériques dans Servan *et al*. 2017).

---

[8] Une approche plus fine, fondée sur un visualisateur des modèles encodeurs/décodeurs (Strobelt *et al*. 2018), qui permet de visualiser les choix lexicaux par probabilité décroissante, confirmerait peut-être que les seuils de probabilité entre les choix lexicaux optimaux sont tels que les substitutions n'opèrent plus à partir de l'époque 13. Le système n'évoluerait plus parce que les alternatives plausibles ont déjà été testées et ne resteraient que des alternatives associées à des probabilités trop faibles.

### *7.2 Une méthodologie pour la portée des segments non-reconnus*

Une première analyse globale des segments non reconnus a mis en lumière l'importance de la ponctuation dans les problèmes de tokenisation. A moyen terme, pour analyser plus finement les sorties, nous souhaiterions distinguer plusieurs niveaux d'analyse : le macro, le méso et le micro. Nous voudrions voir quelles structures sont potentiellement impactées d'une époque à une autre. Le recours à une analyse syntaxique des traductions (*parsing*) pourrait enrichir ce type d'observations « macro ». L'approche méso se centrerait sur l'analyse des changements de catégories grammaticales, ce que la traductologie classique nomme « transposition » (Vinay et Darbelnet, 1966 [1958]). Une étude systématique des étiquettes grammaticales (*pos-tagging*) des traductions nous donnerait ce type d'informations. L'analogie avec la pratique des traducteurs humains est peut-être contestable, mais il nous a semblé qu'un changement portant sur une même catégorie grammaticale (typiquement, le choix d'un synonyme plus judicieux) relevait d'un changement structural moins important. A terme, nous assimilerions cette pratique à une étape sémantique. Ces aménagements à un niveau micro pourraient s'effectuer, par exemple, en utilisant la distance de Levenshtein d'une époque à l'autre entre unités lexicales ayant même catégorie.

## Références

Ballier, N., Amari, N., Merat, L. and Yunès, J.-B. (2020). The Learnability of the Annotated Input in NMT (Replicating Vanmassenhove and Way, 2018 with OpenNMT). *LREC REPROLANG worskhop*.

Britz, D., Le, Q. and Pryzant, R. (2017). Effective Domain Mixing for Neural Machine Translation. *Proceedings of the Second Conference on Machine Translation*, pp. 118-126.

Deng, Y., Kim, J., Klein, G., Kobus, C., Segal N., Servan, C., Wang, B. and Zhang, D. (2017). *SYSTRAN Purely Neural MT Engines for WMT2017*.

Fleury, S. and Zimina, M. (2014). Trameur: A framework for annotated text corpora exploration. In *Proceedings of COLING 2014, System Demonstrations*, pp. 57-61.

Gledhill C., Patin S. and Zimina M. (2017). « Lexico-grammaire et textométrie : identification et visualisation de schémas lexico-grammaticaux caractéristiques dans deux corpus juridiques comparables en français ». *CORPUS 17*.

Klein, G., Kim, Y., Deng, Y., Senellart, J., & Rush, A. M. (2017). OpenNMT: Open-source toolkit for neural machine translation. arXiv preprint arXiv:1701.02810.

Lebart, L. and Salem, A. (1994). *Statistique textuelle*, Paris, Dunod.

Li Z. and Ravela S. (2019). The Learnability of Feedforward Neural Networks in Dissipative Chaotic Dynamical Systems. *AGU fall meeting*, 2019.

Luong, Th., Pham, H. and Manning, C. D. (2015). Effective Approaches to Attention-based Neural Machine Translation, *Proceedings of the 2015 Conference on Empirical Methods in Natural Language Processing*, Lisbon, Portugal, ACL: D15-1166, pp. 1412-1421.

Macé V., Servan C. (2019). Using Whole Document Context in Neural Machine Translation. *16th International Workshop on Spoken Language Translation 2019*, November 2019, Hong-Kong, China.

Montavon, G., Samek, W. and Müller, K-R. (2018). Methods for Interpreting and Understanding Deep Neural Networks. *Digital Signal Processing*. 73, pp.1-15.

Papineni, K., Roukos, S., Ward, T. and Zhu, W. J. (2002). BLEU: a method for automatic evaluation of machine translation. In *Proceedings of the 40th Annual Meeting on Association for Computational Linguistics*, ACL: pp. 311-318.

Post, M. (2018). A Call for Clarity in Reporting {BLEU} Scores. *Proceedings of the Third Conference on Machine Translation: Research Papers*. October 2018, Belgium, Brussels, Association for Computational Linguistics, pp. 186-191.

Salem, A. (1991) « Les séries textuelles chronologiques ». *Histoire & Mesure* 6-1-2, pp. 149-175.

Salkie, R. (1995) INTERSECT: a parallel corpus project at Brighton University. *Computers & Texts 9* (May 1995), pp. 4-5.

Servan, Chr., Crego, J and Senellart, J. (2017) Adaptation incrémentale de modèles de traduction neuronaux. In I. Eshkol and J.-Y. Antoine (Eds.) *24e Conférence sur le Traitement Automatique des Langues Naturelles TALN2017*, pp. 218-226.

Sharma, R.A., Goyal, N., Choudhury, M. and Netrapalli, P. (2018). Learnability of Learned Neural Networks. *ICLR 2018 Conference*.

Strobelt H., Gehrmann S., Behrisch, M., Perer, A., Pfister, H., Rush, AM. (2018). *IEEE transactions on visualization and computer graphics* 25 (1), 353-363, 48. https://github.com/HendrikStrobelt/Seq2Seq-Vis

Sutskever I., Vinyals O. and Le Q. V. (2014). Sequence to sequence learning with neural networks. In *Proceedings of the 27th International Conference on Neural Information Processing Systems*, vol. 2, NIPS'14, pp. 3104-3112, Cambridge, MA, USA: MIT Press.

Tiedemann, J. and Scherrer, Y. (2017). Neural Machine Translation with Extended Context. In *Proceedings of the Third Workshop on Discourse in Machine Translation*, pp. 82-92. Stroudsburg: ACL.

Vinay, J-P. and Darbelnet, J. (1966) [1958]. *Stylistique comparée du français et de l'anglais*, Paris, Didier.

Zhang Y., Lee J., Wainwright M., Jordan M. I. (2017). On the Learnability of Fully-Connected Neural Networks. *Proceedings of the 20th International Conference on Artificial Intelligence and Statistics*, PMLR 54:83-91.

Zimina M. and Fleury S. (2018). Mémoire de traduction (MT) : approche paradigmatique. *Equivalences* 45/1-2, « Des unités de traduction à l'unité de traduction ». (Dir.) C. Balliu, N. Froeliger, L. Hewson, pp. 259-78.